\documentclass[preprint,12pt]{elsarticle}

\usepackage{amssymb}
\usepackage{graphicx}
\usepackage{amsmath}
\usepackage{subfigure}
\usepackage{dblfloatfix}
\usepackage{setspace}
\journal{Materials Discovery}

\begin{document}
\begin{frontmatter}



\title{Building Data-driven Models with Microstructural Images: Generalization and Interpretability}


\author[label1]{Julia Ling}
\address[label1]{Citrine Informatics,\\ Redwood City, CA}
\author[label1]{Maxwell Hutchinson}
\author[label1]{Erin Antono}
\author[label2]{Brian DeCost}
\address[label2]{Materials Science and Engineering Department, \\ Carnegie Mellon University, \\ Pittsburgh, PA}
\author[label2]{Elizabeth A. Holm}
\author[label1]{Bryce Meredig}

\begin{abstract}

As data-driven methods rise in popularity in materials science applications, a key question is how these machine learning models can be used to understand microstructure.  Given the importance of process-structure-property relations throughout materials science, it seems logical that models that can leverage microstructural data would be more capable of predicting property information.  While there have been some recent attempts to use convolutional neural networks to understand microstructural images, these early studies have focused only on which featurizations yield the highest machine learning model accuracy for a single data set.  This paper explores the use of convolutional neural networks for classifying microstructure with a more holistic set of objectives in mind: generalization between data sets, number of features required, and interpretability. 

\end{abstract}


\end{frontmatter}


\section{Introduction}

Recent years have seen a rapid increase in the use of machine learning methods in a variety of materials science applications.  Meredig et al~\cite{Meredig2014} applied data-driven algorithms to density functional theory (DFT) calculations to build a model for stability in ternary compounds.  Sparks et al.~\cite{Sparks2016} demonstrated that machine learning models could be used to screen materials for thermoelectric properties.  Oliynyk et al.~\cite{Oliynyk2016} used machine learning to discover new Heusler alloys.  Xue et al.~\cite{Xue2017} used statistical models and support vector machines to predict transformation temperatures in shape memory alloys.  Collins et al.~\cite{collins2013neural} applied neural networks to predict the properties of titanium alloys given information on the composition and microstructure.  In this case, the microstructure was characterized by several key metrics that were pre-calculated from the SEM image.  Machine learning methods have also been used in discovering materials for organic light emitting diodes~\cite{Bombarelli2016}, in finding new piezoelectric materials~\cite{Xue2016}, and in classifying polymer states~\cite{Wei2017}.   The promising results in these and many other studies demonstrate the potential for machine learning to fundamentally change how new materials are discovered and how processing steps are optimized.

A key challenge in applying machine learning algorithms to materials science data is that materials science data comes in many forms.  Determining how to featurize different types of data so that they can be used as inputs to machine learning algorithms is not always straightforward.  Some data, such as yield stresses at room temperature, come in scalar form and can be used directly as inputs to machine learning models.  Other data, such as X-ray diffraction patterns, require more specialized analytics~\cite{Iwasaki2017}.  

One type of data that is ubiquitous in materials science applications is microstructural images.  Because of the known dependence of macroscopic properties on material microstructure, scanning electron microscope (SEM) images are a commonly collected data type for a wide range of different material categories.   However, using these SEMs in data-driven methods is not straightforward because the data come in the form of images, not scalar data.  These images can be represented as 2-dimensional or 3-dimensional (in the case of color images) arrays.  The convolutional neural network (CNN) is a machine learning algorithm that was designed specifically to leverage the correlative structure in images to make predictions on image data.  CNNs have been widely applied to problems such as image classification~\cite{Ciresan2011, Krizhevsky2012}, facial recognition~\cite{Lawrence1997, Parkhi2015}, and natural language processing~\cite{Hu2014}. 

Convolutional neural networks have also been applied to the problem of SEM classification.  DeCost et al.~\cite{DeCost2017} applied CNNs to classify SEMs of steels based on their primary microconstituents.  They used a pre-trained neural network and applied Vector of Locally Aggregated Descriptors (VLAD) encoding to the outputs of the final convolutional layer.  In their method, the CNN was used to calculate image-based features, which were then fed into a support vector machine (SVM) which performed the final classification.  They noted that using the CNN to calculate the image-based features improved classification accuracy versus applying VLAD to the raw image data.  A noteworthy trait of their workflow is that the VLAD encoding leveraged the translation-invariant nature of SEMs: texture is more important than where a given object occurs in the image.  These SEM features have the potential to be used alongside processing data and experimental measurements to construct data-driven process-structure-property maps.   

Others have also applied neural networks to SEM featurization.  Azimi et al.~\cite{Azimi2017} used a fully connected neural network to make pixel-by-pixel classifications of phase in steel SEM images.  Kondo et al.~\cite{Kondo} used a convolutional neural network to featurize a SEM data set for ceramic materials and demonstrated that these methods could be effective for small data sets.  These studies demonstrate that CNNs can be used effectively to featurize microstructural images.  However, all three of these studies evaluated their methods over a single data set.  For CNN-based featurization of SEM images to gain widespread use, it will be important to have featurizations that are both generalizable and scalable.  The same featurization should work across multiple data sets without data set-specific hand-tuning.  This paper will evaluate different CNN-based featurizations of SEMs over multiple data sets to explore the generalizability and interpretability of these methods.  

Section~\ref{MethodologySec} will provide some background on CNNs in general, as well as the specific implementation and featurizations used in this study.  Section~\ref{TestCasesSec} will introduce the data sets on which these featurizations were evaluated.  The results of this evaluation, as well as an investigation of the interpretability of CNN-based featurizations, is presented in Section~\ref{ResultsSec}.  Finally, conclusions and next steps are discussed in Section~\ref{ConclusionsSec}. 

\section{Machine Learning Methodology} \label{MethodologySec}
\subsection{Background on Convolutional Neural Networks}
CNNs are a type of neural network that were designed specifically to process image data~\cite{Krizhevsky}.  They consist of stacks of layers that apply 2D convolutional filters to the image to detect the presence or absence of various patterns.  While the earlier layers tend to detect simpler patterns, such as the presence of edges, later layers detect higher-level patterns, such as the presence of faces in the image.  These networks have been successfully applied to a wide variety of image classification tasks~\cite{Krizhevsky, Lawrence}.  

\begin{figure*}[t]
\begin{center}
\includegraphics [scale=.85]{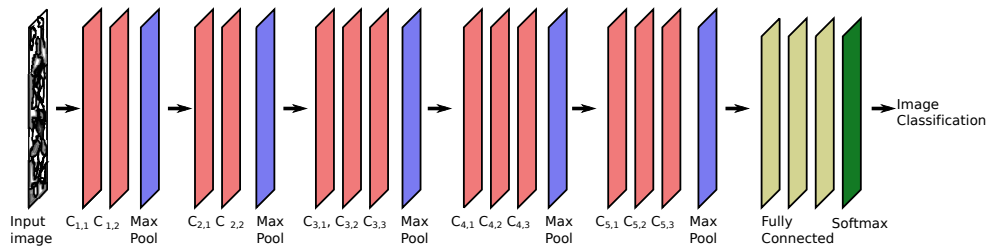}
\end{center}
\caption{Schematic of the VGG16 CNN architecture.  There are a series of 5 stacks of convolutional layers.  In each stack, there are two or three convolutional layers followed by a max pooling layer.  The convolutional layers are indexed based on their position within a stack (e.g. $C_{1,2}$ refers to the second convolutional layer in the first stack of layers).  Finally, there are three fully connected layers which provide a final predicted classification.} 
\label{vgg_fig} 
\end{figure*}

Figure~\ref{vgg_fig} shows the architecture of one such widely used CNN, called VGG16~\cite{Simonyan}.  In this architecture, there are five stacks of convolutional layers, each consisting of two or three convolutional layers followed by a max pooling layer.  The max pooling layers reduce the dimensionality of the output by taking only the maximum activation over a set of outputs.  Finally, there are three fully connected layers connected to the softmax output layer that renders the final prediction, which in the case of VGG16 includes 1000 different categories of objects commonly found in photographs, such as bell pepper, zebra, volcano, stove, and scuba diver.  

CNNs typically require a large corpus of images to train.  For example, the ImageNet database contains over 14 million labeled images~\cite{Russakovsky}.  These images are used to determine the optimal convolutional filters for each convolutional layer, as well as the optimal weights for the fully connected layers.  The number and size of filters per layer are pre-specified as part of the CNN architecture.  While it can be a daunting task to find a labeled data set of millions of images specific to a given classification problem, it is usually possible to use transfer learning to sidestep this obstacle.  In transfer learning, the filters and weights from a neural network that was fully trained for some other task are used for a new task.  With this initialization, it is often possible to achieve good results with a much smaller data set of images for training~\cite{Shin}.  Often, the filters are left unchanged between different classification tasks and only the fully connected layers are re-trained to the new data set.  

\subsection{Implementation}

\begin{figure}[t]
\begin{center}
\includegraphics [scale=.4]{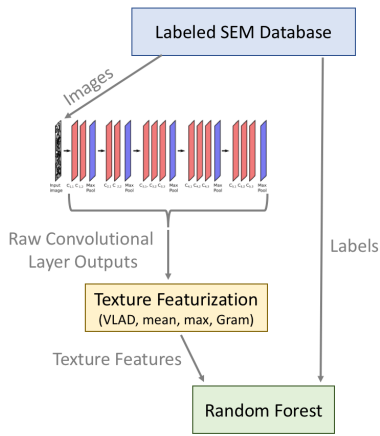}
\end{center}
\caption{Schematic of the featurization and classification workflow.} 
\label{workflow_fig} 
\end{figure}

There were three steps in the feature evaluation workflow: i) transformation of the image by the CNN, ii) texture featurization of the outputs of the CNN, and iii) random forest classification using the texture features as inputs.  These steps are shown schematically in Fig.~\ref{workflow_fig} and described in more detail in the following subsections.

\subsubsection{Transformation by CNN}

For the purposes of this study, the pre-trained VGG16 neural network was used.  It was implemented using Keras~\cite{chollet2015keras}, with a Tensorflow backend.  No network re-training was performed--the pre-trained convolutional filters from VGG16 were used directly.  The fully connected layers were not used at all, since those are needed only for the original classification task (i.e. zebra vs stove), not for image featurization.  In this study, the CNN was not used for classification, just for image featurization.  In other words, the network was used to convert the raw pixel data into a tensor format that contained information about what patterns were present in the image.  

One question of interest is from which convolutional layer the features should be extracted.  The output from earlier convolutional layers will contain information about lower level features and simpler textures.  The output from later layers will contain information about higher level features and more complex textures.  DeCost et al.~\cite{DeCost2017} compared the outputs of the $C_{4,3}$ and $C_{5,3}$ layers and saw little difference in performance on their data set, but did not compare to the outputs from the earlier convolutional layers.  In Section~\ref{ResultsSec}, we explore the effect of the choice of convolutional layer on the accuracy of the models built on those feature sets.

\subsubsection{Texture Featurization}

\begin{table*}[b]
\centering
\caption{Number of features produced by each featurization technique for each VGG Layer}\label{featurization_table}
\begin{tabular}{| l |  l | l | l | l | l|}
\hline
 \textbf{Featurization Strategy} & $\bf{C_{1,2}}$ &$\bf{C_{2,2}}$ & $\bf{C_{3,3}}$ & $\bf{C_{4,3}}$ & $\bf{C_{5,3}}$ \\\hline
 VLAD & 2048 & 4096 & 8192 & 16384 & 16384 \\\hline
 Mean & 64 & 128 & 256 & 512 & 512 \\\hline
 Max & 64 & 128 & 256 & 512 & 512 \\\hline
 Gram & 4096 & 16384 & 65536 & 262144 & 262114 \\\hline
 \end{tabular}
\end{table*}

In the context of SEM microstructural images, we care about texture, not about the presence or absence of distinct objects like zebras or stoves.  Therefore, the outputs from the convolutional layers were processed in such a way as to extract these texture features.  In particular, the location of a given feature within the SEM image is not important, so the featurizations should be translation invariant.  Four different translation-invariant texture featurization strategies were compared.  The number of features generated with each of these strategies using different VGG output layers is presented in Table~\ref{featurization_table}.

\begin{itemize}
  \item \textbf{VLAD} The Vector of Locally Aggregated Descriptors (VLAD) encoding~\cite{vlad} was used by DeCost et al.~\cite{DeCost2017}.  In this featurization, the outputs from a given convolutional layer are clustered and the cluster centroids are treated as a dictionary of visual words.  In VLAD, any new image is featurized by first passing it through the CNN, then taking the difference between the CNN convolutional layer output and the dictionary words.  These residuals form a feature vector of length $N_{filters} \cdot N_{words}$, where $N_{filters}$ is the number of filters in the output convolutional layer and $N_{words}$ is the number of centroids specified during the k-means clustering phase of VLAD featurization.  $N_{words}$ was set to 32 in this study, in keeping with Ref.~\cite{DeCost2017}.  
  
  One downside of this featurization strategy is that it requires forming a dictionary for a given data set.  If the data set is updated with new images, the dictionary should be recalculated.  Furthermore, the process of forming this dictionary can also be memory intensive; to reduce this memory burden, the images were first rescaled to be 255 by 255 pixels before this featurization method was applied.  

  More details on this featurization are available in Refs.~\cite{vlad, DeCost2017}.  
  
    \item \textbf{Mean} In this very simple post-processing step, the output from a given convolutional layer was averaged spatially, so that the resulting feature vector was of length $N_{filters}$, the number of filters in the output convolutional layer.  In other words, the feature set was given by $ mean_{j}[ F_{ij}]$ where $F_{ij}$ represents the output of a specified convolutional layer.  The index $i$ refers to the filter number and $j$ refers to the spatial location. 
  \item \textbf{Max} In this method, the maximum value of the output from a given convolutional layer was taken, $ max_{j}[ F_{ij}]$.  The resulting feature vector was length $N_{filters}$.
    \item \textbf{Gram} The Gram matrix is given by $G_{ij} = \sum_{k} F_{ik}F_{jk}$.  The Gram matrix has been previously used to assess texture in the context of artistic styles in artwork~\cite{Gatys}.  Lubbers et al.~\cite{lubbers2016inferring} applied this featurization to unsupervised learning on SEM images. This method results in $N_{filters}^{2}$ features.  Of the featurization strategies explored, it therefore results in the largest number of generated features, which can be disadvantageous because it can lead to over-fitting and increased computational expense when using these features to train a data-driven model.

\end{itemize}

\subsubsection{Random Forest Classification}

The texture vectors generated by the featurization strategies were used as inputs to a random forest classifier, a computationally efficient machine learning algorithm that does not require as large a training set as neural networks in order to have good accuracy.  The random forest classifier had 400 trees in the ensemble and was evaluated with 10 trials of 3-fold cross-validation.  The mean performance and variance in that performance were assessed over the 10 trials.  The performance was evaluated via the F1 score, an accuracy metric for multi-class classification that has a value of 1.0 for a perfect classifier and 0.0 for a classifier that is always wrong.

\section{Test Data Sets} \label{TestCasesSec}

The various featurizations were compared on three different SEM data sets: a data set of titanium alloys with varying micro-textured regions, a data set of steels processed with varying heat treatments, and a data set of synthetic SEM images of powder materials with different particle size distributions.  These data sets are all available on Citrination\footnote{Titanium: https://citrination.com/datasets/154195/} \footnote{Steel: https://citrination.com/datasets/152980/}  \footnote{Powder: https://citrination.com/datasets/154196/}.  Table~\ref{dataset_table} presents a summary of the number of categories and images in each data set.  In all cases, some basic image pre-processing was performed before featurization.  This pre-processing included cropping off the scale bars and legends, which could provide false signal to the neural network.  

\begin{table*}[b]
\centering
\caption{Summary of Test Data Sets}\label{dataset_table}
\begin{tabular}{| l |  p{3.3cm} | p{2.5cm}|}
\hline
 \textbf{Data Set} & \textbf{\# of categories} & \textbf{\# of images} \\\hline
 Titanium & 4 & 60 \\\hline
 Steel & 3 & 710 \\\hline
 Powder & 8 & 2048 \\\hline
\end{tabular}
\end{table*}

\subsection{Titanium Data set}

\begin{figure}[t]
\begin{center}
\subfigure[Billet]{\label{} \includegraphics[width=40mm]{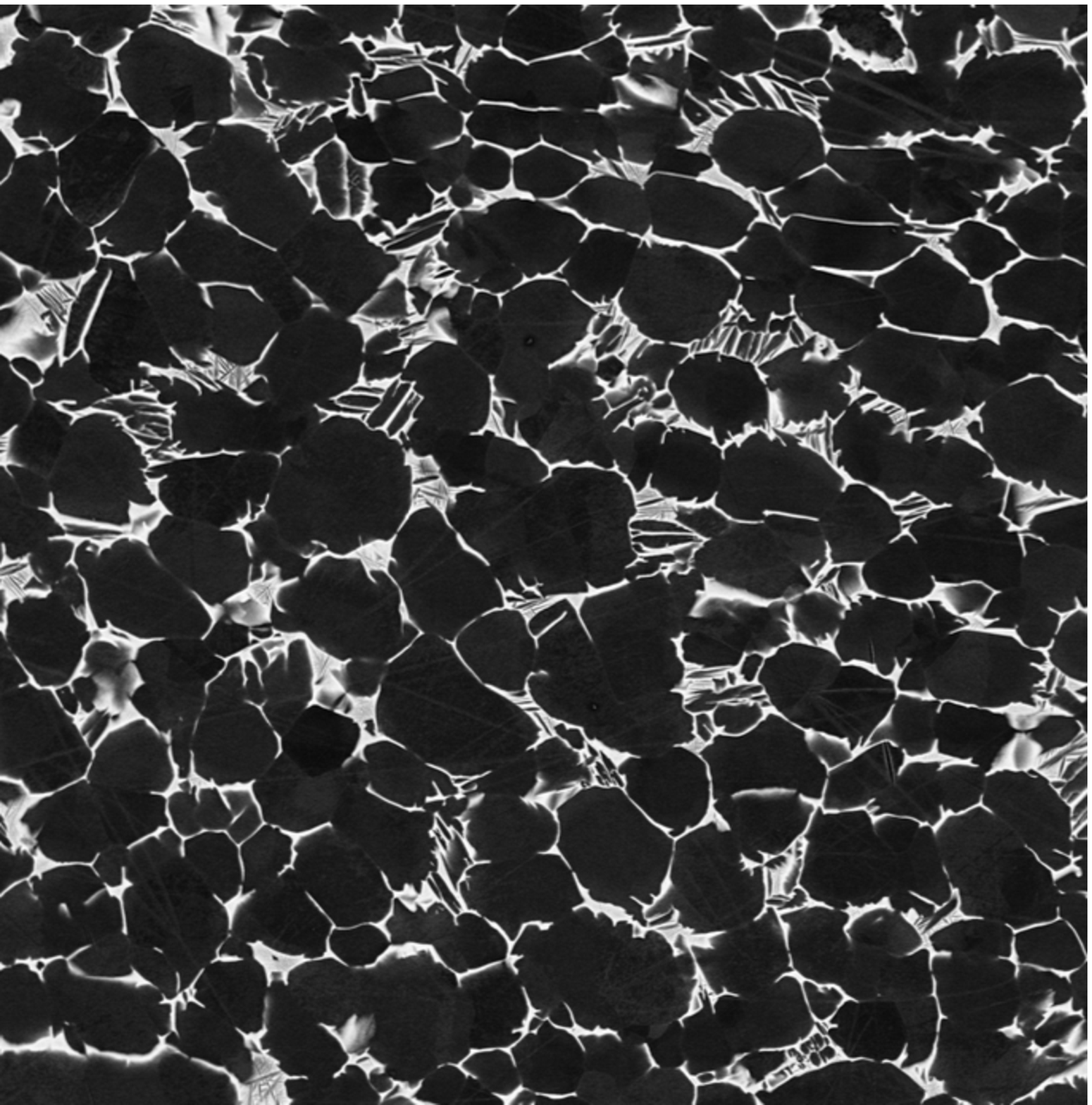}}
\subfigure[Forged $0^{\circ}$]{\label{} \includegraphics[width=40mm]{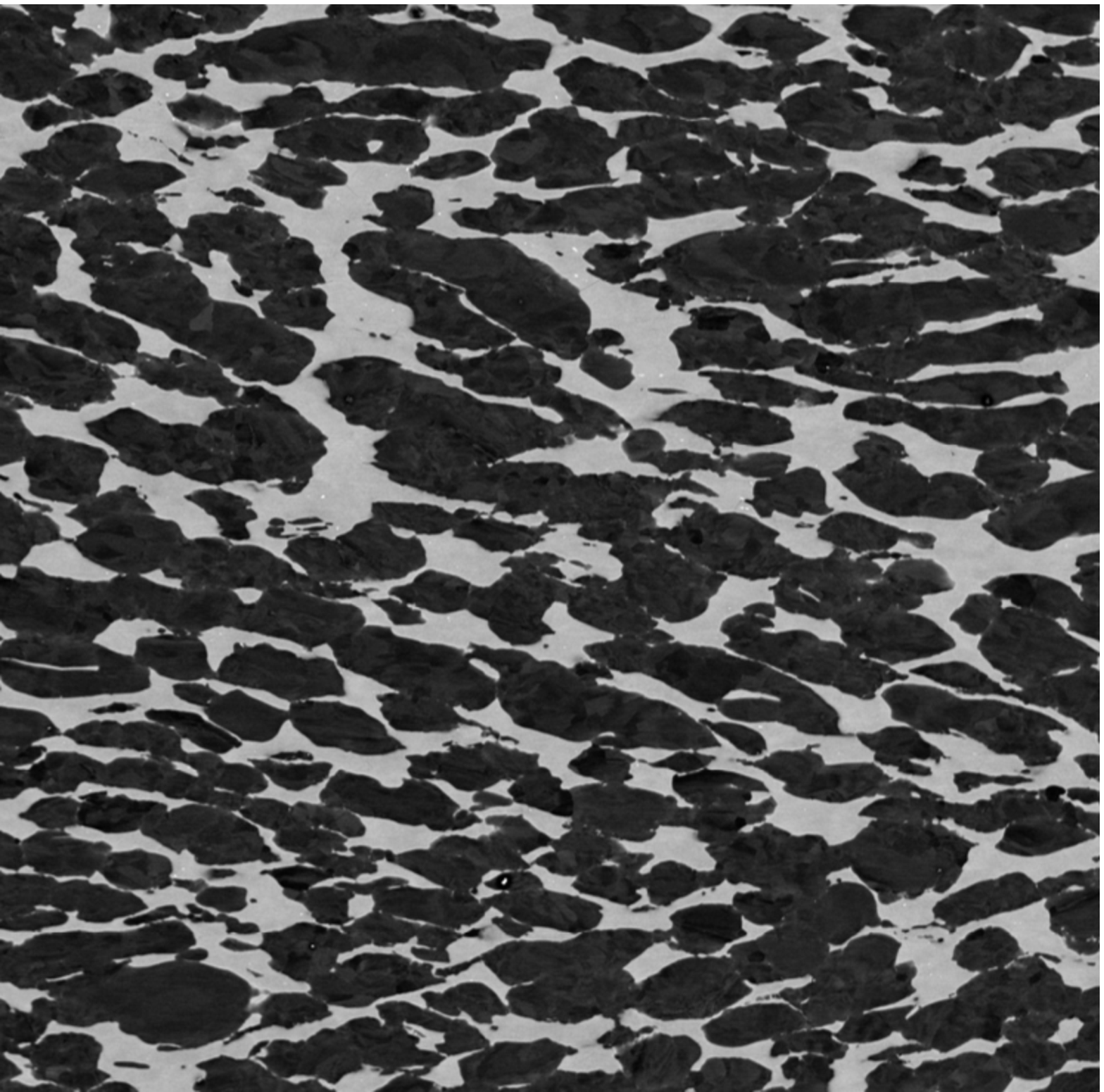}}
\subfigure[Forged $45^{\circ}$]{\label{} \includegraphics[width=40mm]{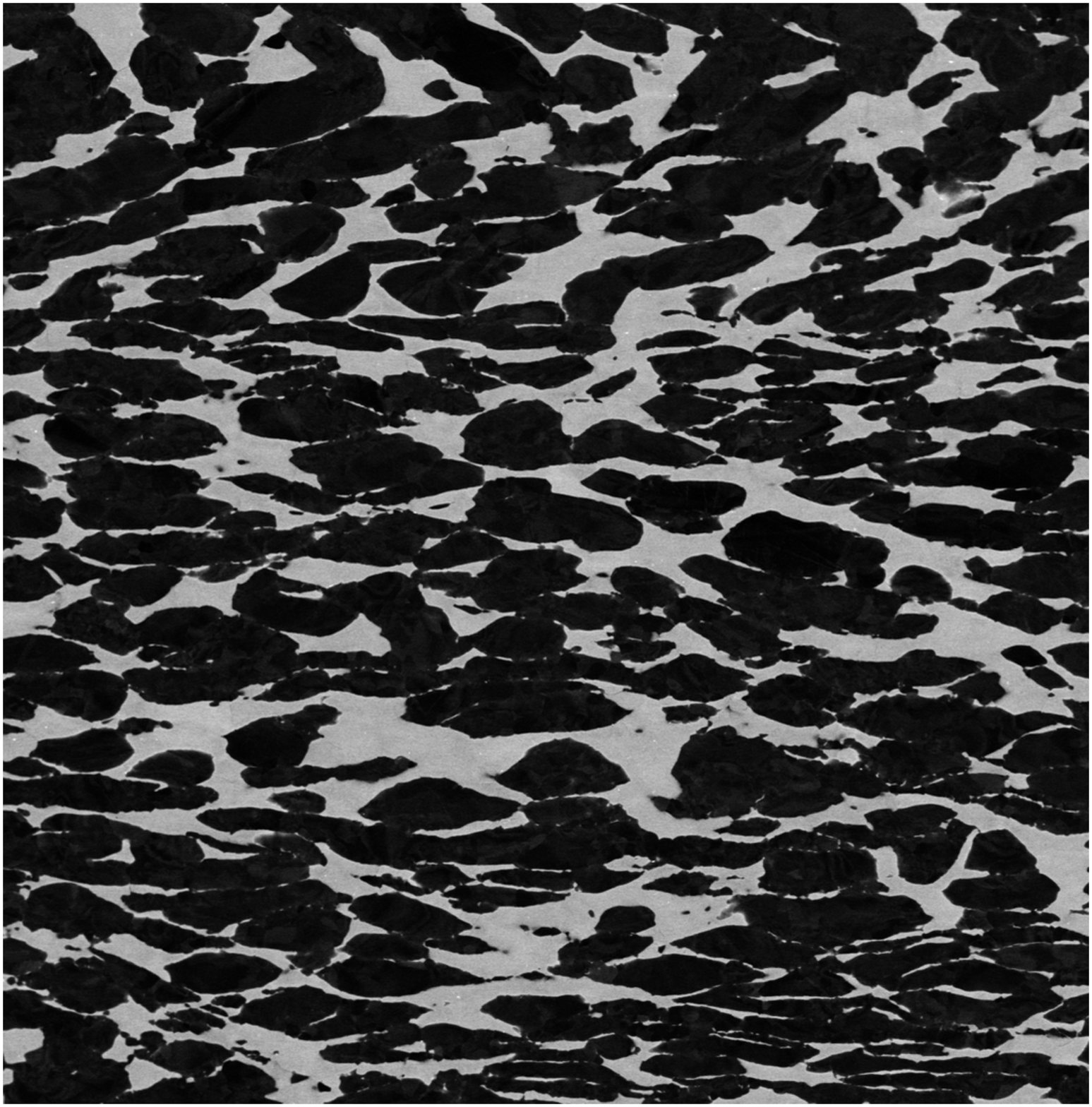}}
\subfigure[Forged $90^{\circ}$]{\label{} \includegraphics[width=40mm]{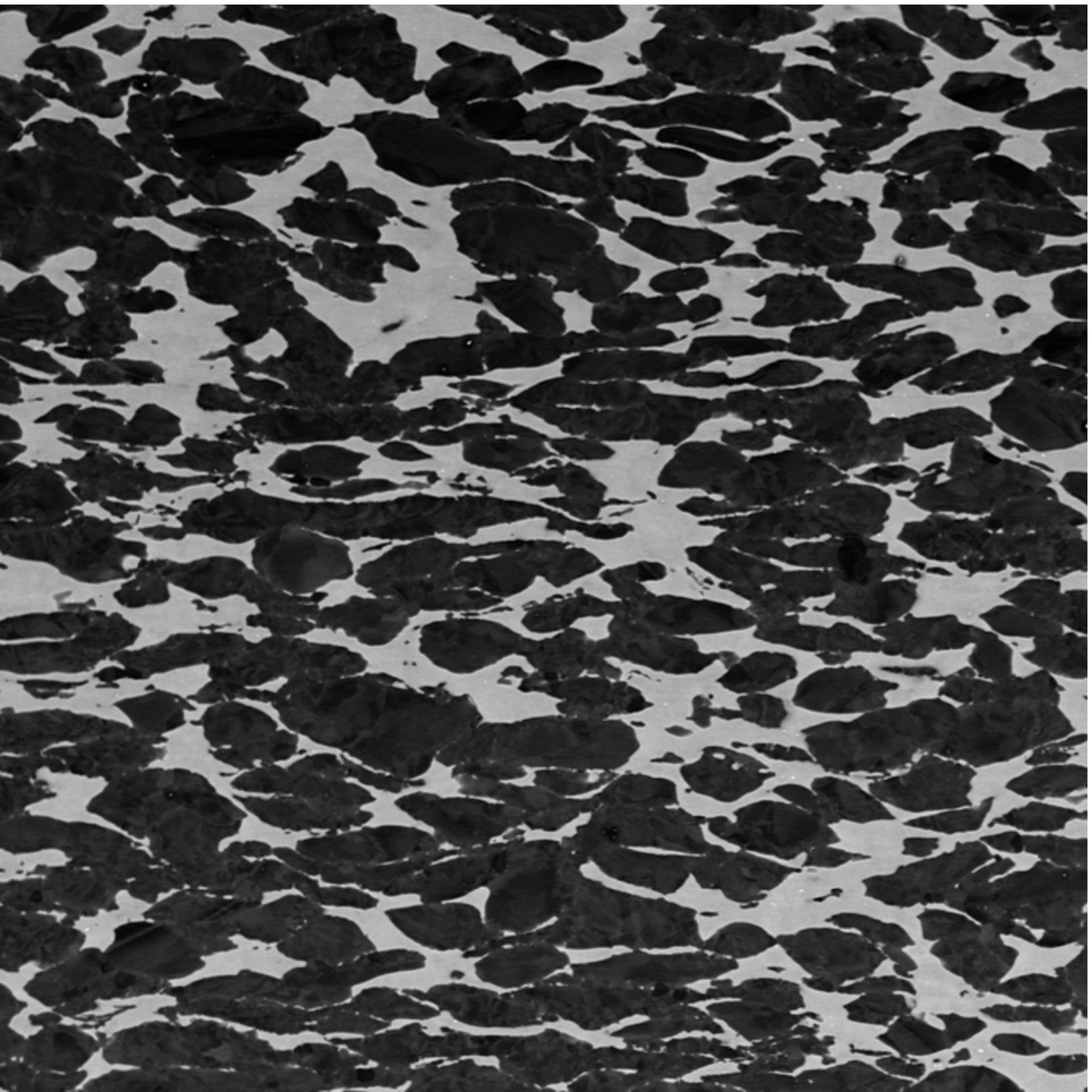}}
\caption{Examples of SEM images from the titanium data set.}
\label{Ti_sem_fig}
\end{center}
\end{figure}

The titanium data set was from Pilchak et al.~\cite{pilchak2016} and was composed of SEM images of the near-alpha Ti-6Al-2Sn-5Zr-2Mo-0.1Si alloy.  These images were taken of the billet material as well as of forged samples with different angles ($0^{\circ}$, $45^{\circ}$,$90^{\circ}$) between the microtextured regions and the direction of metal flow during manufacture.  There were 60 SEM images in all, 15 per category.  The machine learning task was to differentiate between four categories: the billet and the three different forging directions.  Sample SEMs from these four categories are shown in Fig.~\ref{Ti_sem_fig}.  This data set represents a challenging test case, as these categories are difficult to differentiate visually.  The images from this data set were rescaled from over 2000 pixels per side to a size of 512 by 512 pixels before applying the CNN transformation.  While the convolutional layers can be applied directly to images of any size, the filters are not optimal for images of all sizes.

\subsection{Steel Data set}

\begin{figure}[t]
\begin{center}
\subfigure[Network]{\label{} \includegraphics[width=50mm]{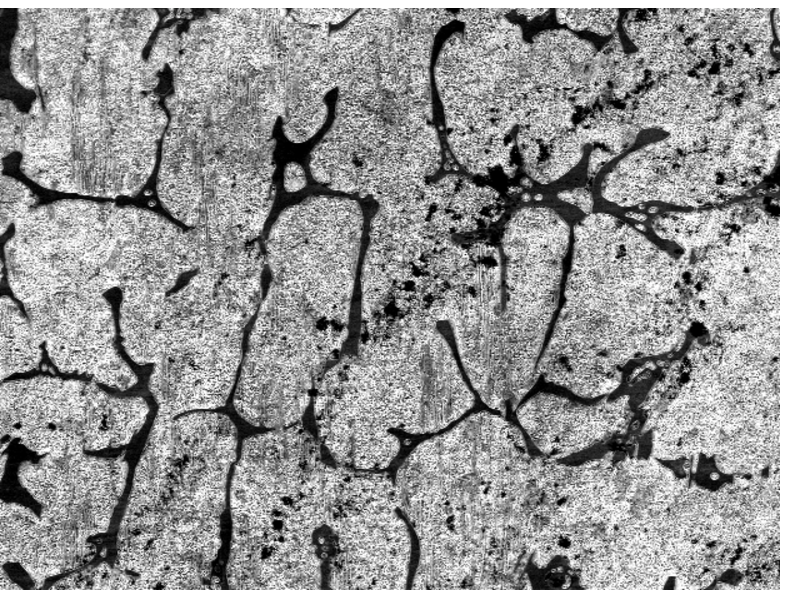}}
\subfigure[Spheroidite]{\label{} \includegraphics[width=50mm]{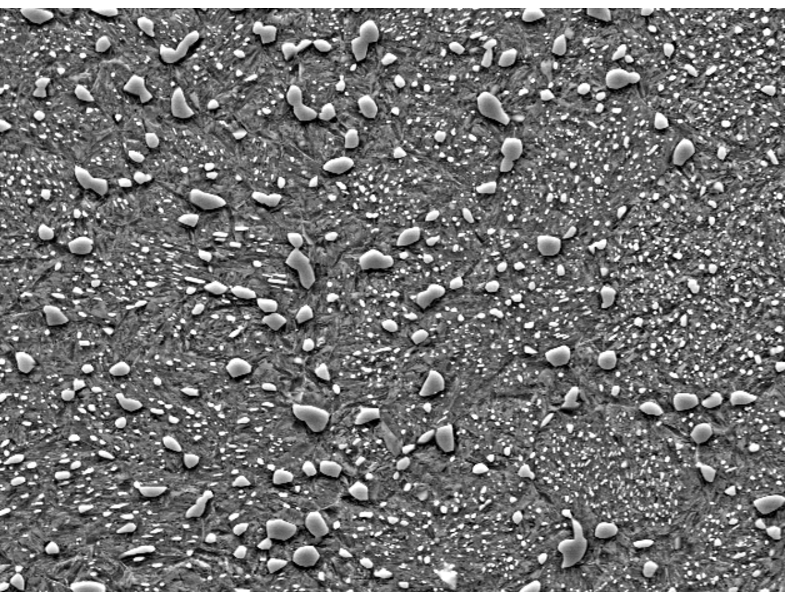}}
\subfigure[Pearlite]{\label{} \includegraphics[width=50mm]{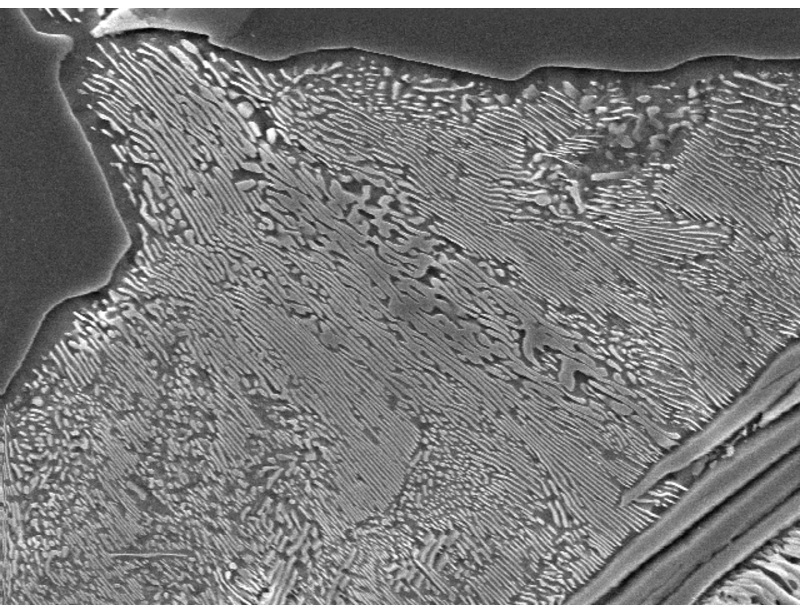}} 
\caption{Examples of SEM images from the steel data set}
\label{steel_sem_fig}
\end{center}
\end{figure}

The steel data set was from DeCost et al.~\cite{UHCSDB}.  It consisted of 710 images of ultra high carbon steels subjected to different annealing treatments and cooling treatments.  These images were labeled as belonging to one of three categories based on their primary microconstituent: network, spheroidite or pearlite.  Figure~\ref{steel_sem_fig} shows a sample SEM from each of these categories.  The task for the random forest classifier was predicting this primary microconstituent given the input SEM image.

\subsection{Powder Data set}

\begin{figure}[t]
\begin{center}
\subfigure[Distribution 1]{\label{} \includegraphics[width=40mm]{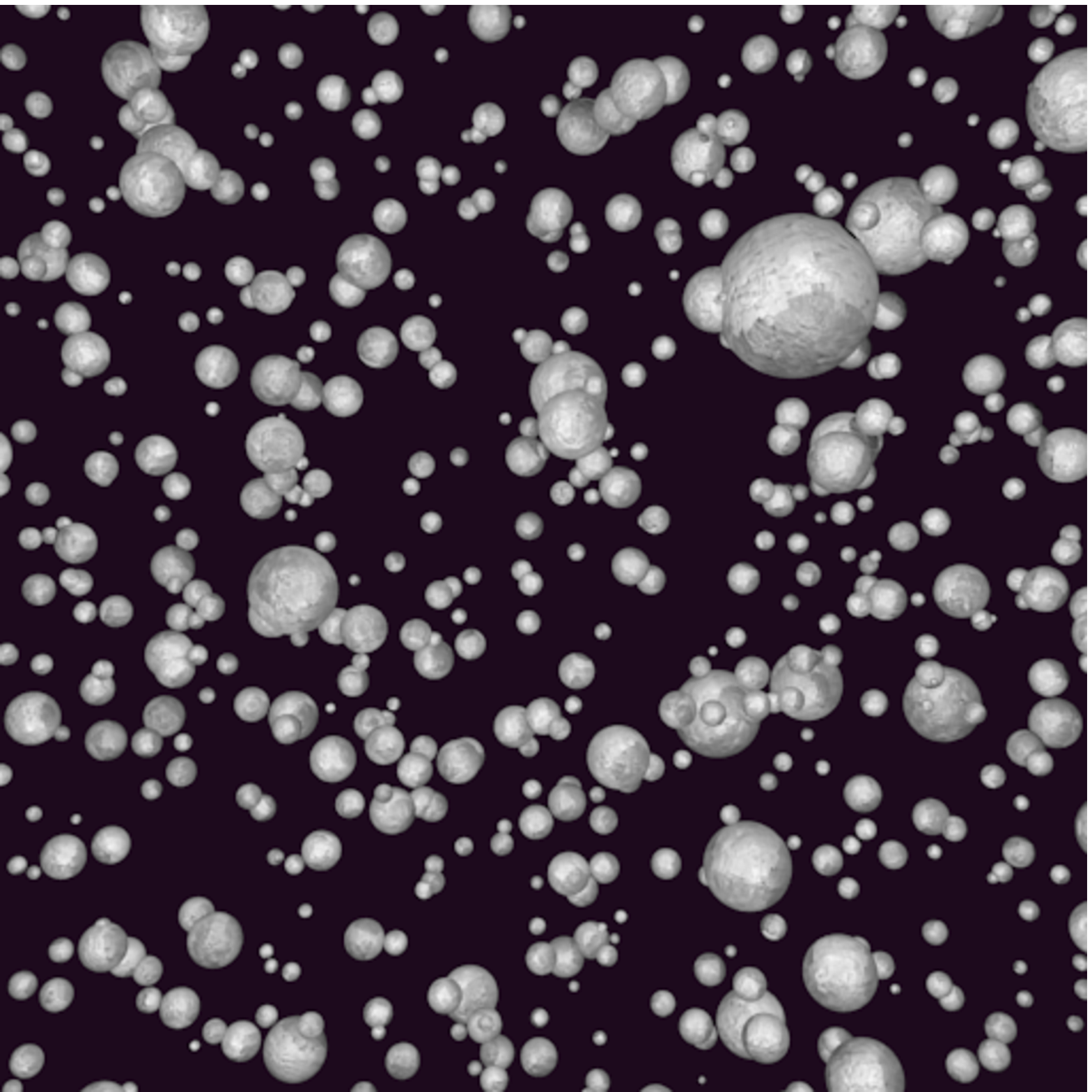}}
\subfigure[Distribution 2]{\label{} \includegraphics[width=40mm]{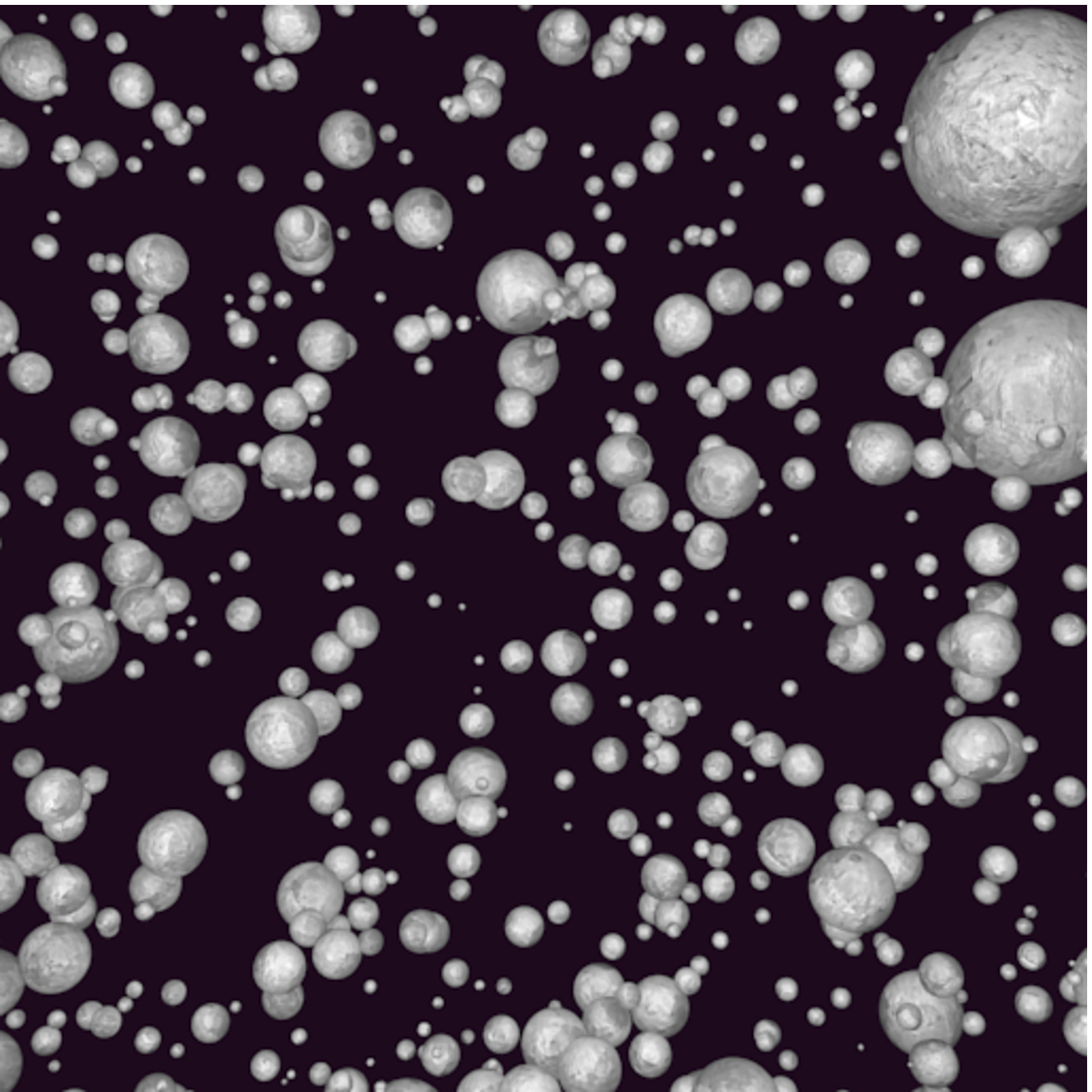}}
\caption{Examples of SEM images from the powder data set.}
\label{am_sem_fig}
\end{center}
\end{figure}

This data set was from Ref.~\cite{DecostAM}, and included 2048 images of synthetic SEMs of powder materials relevant to additive manufacturing applications.  There are 8 different particle size distributions, with 256 images representing each distribution.  In this case, the random forest classifier was trained to classify the 8 different distributions based on the SEM images.  Figure~\ref{am_sem_fig} shows sample SEMs from two of the distributions.  As this figure shows, these SEMs are difficult to differentiate via visual inspection.  Because of the larger size of this data set, the VLAD dictionary was formed using only 20\% of the total number of images to reduce the required memory for this featurization process.

\section{Results} \label{ResultsSec}
\subsection{Model Performance}
For each of the three data sets, we varied two parameters: i) the CNN layer from which to take the outputs and ii) the texture featurization strategy.  The outputs from five different CNN layers were evaluated: $C_{1,2}$, $C_{2,2}$, $C_{3,3}$, $C_{4,3}$, $C_{5,3}$.  These five layers have 64, 128, 256, 512, and 512 filters respectively.  The outputs from these layers were then featurized in four different ways, Gram, max, mean, and VLAD, as described in Section~\ref{MethodologySec}.  In all, then, 60 different feature sets were generated: 3 data sets, 5 output layers, and 4 featurizations.  

\begin{figure*}[t]
\begin{center}
\includegraphics [width=150mm]{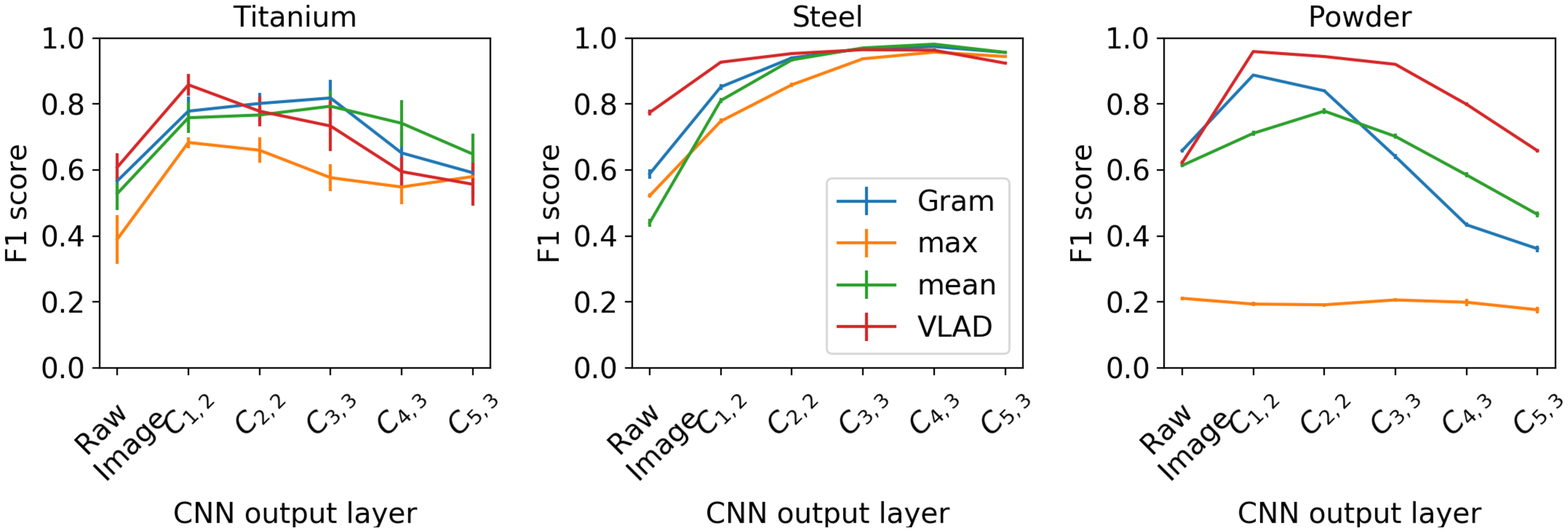}
\end{center}
\caption{Model accuracy of random forests using input feature sets generated via 4 different texture featurizations and 5 different CNN layers.  The accuracy of these featurization techniques on the untransformed input image is also shown for comparison.} 
\label{f1_fig} 
\end{figure*}

Figure~\ref{f1_fig} shows the F1 score for each of these 60 cases.  In addition, the F1 scores for the featurizations applied to the untransformed image are also presented.  The error bars are given by the standard deviation in the F1 score over the 10 trials.  The performance of the VLAD featurization strategy on the steel test case is comparable to that reported by Decost et al.~\cite{DeCost2017} for this same case.  Similarly, the performance of VLAD on the powder data set was also similar to that reported by DeCost et al.~\cite{decost2017characterizing} for that data set.  Notably, DeCost et al. used a different classification algorithm (Support Vector Machines), suggesting that the classification algorithm is less important than the featurization method.  For all three of these cases, relatively accurate models ($F1 > 0.8$) were created.  This is noteworthy since the classification tasks for the titanium and powder cases were visually challenging.

 As Fig.~\ref{f1_fig} shows, there is no single texture featurization and CNN layer combination that is optimal for all three data sets.  The max texture featurization performs poorly for all three data sets.  Similarly, the untransformed input image is not optimal for any data set--it is always preferable to use at least one CNN block to filter the image.  The other three texture featurizations are comparable: VLAD does better for the powder data set, but slightly worse for the steel data set than the Gram and mean featurizations.  The Gram and mean featurizations have similar performance to each other for all three data sets.  These results suggest that the mean featurization, which is simple to implement, computationally inexpensive, and comparatively memory efficient (since it requires only $N_{filter}$ features) is a reasonable choice for texture featurization.

The influence of CNN output layer is also striking.  We see that the model performance is best for later layers in the steel case, best for earlier layers in the powder case, and best for the middle layers in the titanium case.  Notably, there does not seem to be any correlation between data set size and which output layer should be used.   It is possible that the layers later in the CNN are more relevant to the steel data set than the powder data set because the steel textures, with lamellar and cell-like structures, are more complex than the powder textures, which consist mainly of intersecting curves and circles.  These results demonstrate that the optimal output layer depends on the test case of interest and there can be significant degradation in performance for choosing the wrong layer.

\begin{figure*}[t]
\begin{center}
\includegraphics [scale=.5]{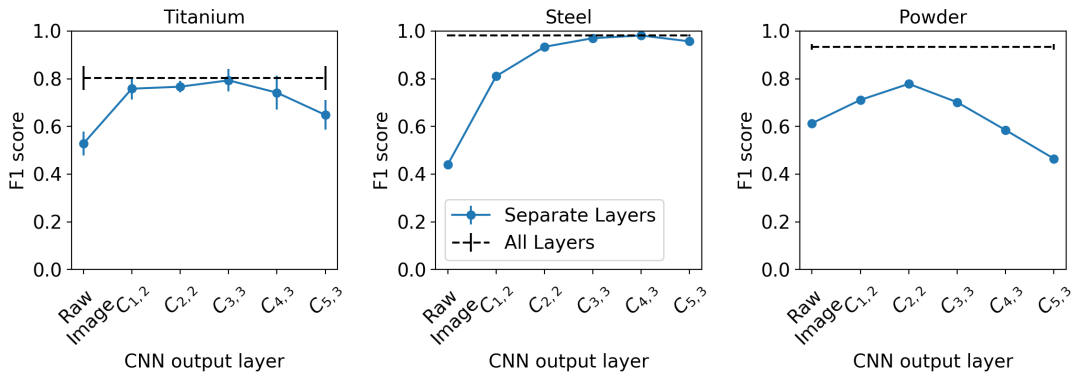}
\end{center}
\caption{Model accuracy of random forests using feature sets generated using the mean texture strategy from outputs of five different convolutional layers separately and together.} 
\label{all_layers_fig} 
\end{figure*}

Based on these results, it was of interest to evaluate the model accuracy when using outputs from all five of these layers.  Fig.~\ref{all_layers_fig} shows the random forest model accuracy with the mean texture featurization using the outputs from each of the layers separately versus using the outputs from all of the layers together.  As this figure shows, using the features generated from all five of the convolutional layers provided at least as accurate a random forest model as using any of the single layer feature sets individually.  In the powder case, the performance using all five layers was significantly higher than when using only individual layers.  It seems possible that the combined feature set is most beneficial when working with larger data set sizes, since the risk of overfitting is less strong for larger data sets.  

\subsection{Texture Visualization}
One of the other benefits of using the mean texture featurization is that these textures can be visualized.  The VLAD featurization is based on combinations of filter outputs which form a dictionary of words.  The Gram featurization is based on correlations between pairs of filters.  In the mean texture featurization, on the other hand, each texture is based on a single convolutional filter.   It is therefore straightforward to run an optimization routine to determine which input image would cause that convolutional filter to be most strongly activated~\cite{Zeiler}.  

To run this optimization, we made use of code that was provided as part of the Keras package~\cite{chollet2015keras}.  An image with random white noise is run through the VGG network, and the output of a given convolutional filter is recorded.  Using gradient ascent, the mean activation of that filter over the image is maximized by iteratively modifying the input image.  The input image that maximizes the mean value of the filter activation depicts the texture represented by that filter.

\begin{figure}[t]
\begin{center}
\subfigure[$C_{1,2}$]{\label{} \includegraphics[width=30mm]{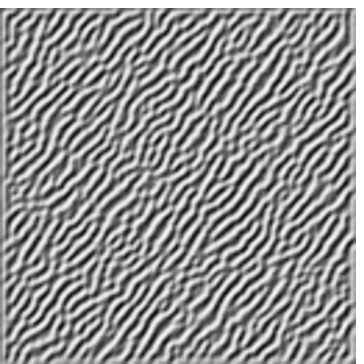}}
\subfigure[$C_{2,2}$]{\label{} \includegraphics[width=30mm]{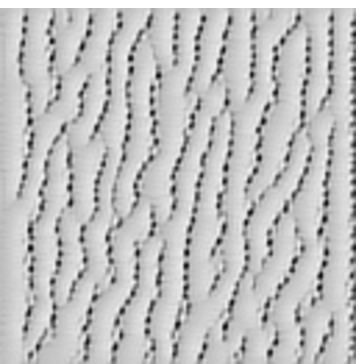}}
\subfigure[$C_{3,3}$]{\label{} \includegraphics[width=30mm]{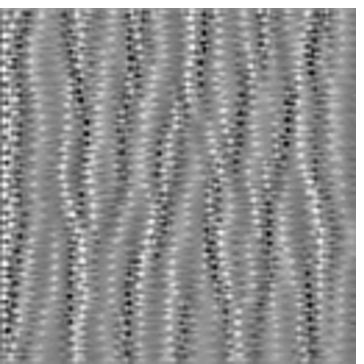}}\\
\subfigure[$C_{4,3}$]{\label{} \includegraphics[width=30mm]{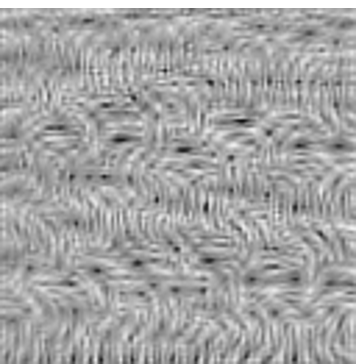}}
\subfigure[$C_{5,3}$]{\label{} \includegraphics[width=30mm]{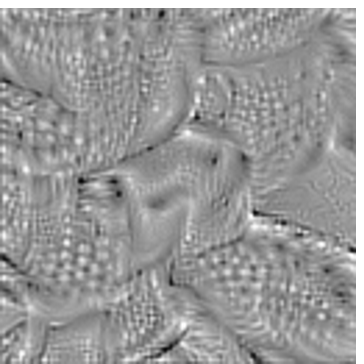}}
\caption{Sample textures from different convolutional layers.}
\label{texture_layers_fig}
\end{center}
\end{figure}

Figure~\ref{texture_layers_fig} shows sample textures from each of the 5 layers.  As this figure illustrates, the textures represented by the earlier layers are more local, whereas the textures in the later layers are more hierarchical and complex.  These textures are not specific to any given test data set; they are based on the pre-trained filters of VGG16.

\begin{figure*}[t]
\begin{center}
\includegraphics [width=100mm]{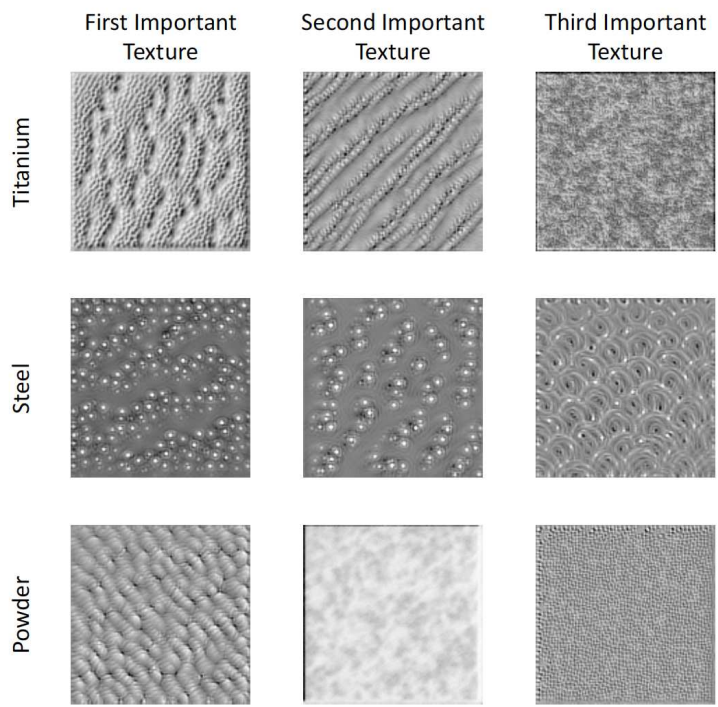}
\end{center}
\caption{The most important textures for the random forest model for each data set.  From left to right, the columns represent the first, second, and third most important texture features for each case.  From top to bottom, each row represents a different case: titanium, steel, and powder.}
\label{texture_fi_fig} 
\end{figure*}

Random forest models can provide a sense of which input features are most important to a given classification task.  Those features that are used most often in splitting criteria and which give the greatest reduction in node impurity are more important than features that get used less often in splits.  It is interesting to examine which texture features were most important for each test case.  Fig.~\ref{texture_fi_fig} shows the three most important textures for each of the three test cases, as assessed via random forest feature importance.  

These textures are interesting to visualize because they give some indication as to how the random forest uses the CNN texture features to differentiate between microstructures.  For example, it is not surprising that in the steel case, two of the important textures are dotted, since one of the categories in that case was spheroidite.  The mean value of those three texture features over all the images with the spheroidite label is [559, 329, 33].  The mean value over those three texture features for all the images with the network label is [72, 55, 3].  Notably, these values are unitless--they are simply measures of intensity, where higher values indicate that a given texture is more present in the data set.  The very high values of intensity for the first two textures in the spheroidite case make sense because those textures have dots that mimic the spheroidite microstructure.  Because these two textures have high intensity in many of the spheroidite images, but much lower intensity in the pearlite and network images, these textures are very useful to the random forest in determining which microconstituent is present in a given image.  The fact that these textures are the most important in this case indicates that the classifier prioritized the task of distinguishing between spheroidite and non-spheroidite SEMs.  This prioritization is not surprising given that the majority (53\%) of the SEMs in this case were in the spheroidite category. 

The relevance of the important textures in the other two cases is less visually obvious.   In the powder test case, the third most important texture has many small adjoining dimples, a reasonable texture to use to distinguish between particle size distributions.  

\begin{figure*}[t]
\begin{center}
\includegraphics [width=80mm]{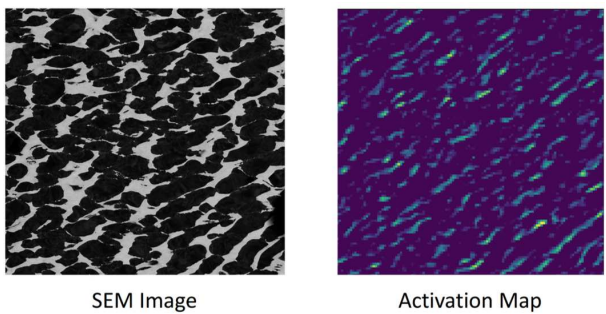}
\end{center}
\caption{The figure on the left shows a sample SEM image from the Titanium case, from the Forged $0^{\circ}$ category.  The heat map on the right shows the activation of the convolutional filter associated with the second important texture for this case.}
\label{activation_fig} 
\end{figure*}

In the titanium case, the second texture is less intense on average in the $90^{\circ}$ forged and billet categories than in the other two categories.  Looking at the sample SEM images in Fig.~\ref{Ti_sem_fig}, it seems that the $90^{\circ}$ forged and billet SEMs have fewer diagonal structures like those represented by this texture.  Figure~\ref{activation_fig} shows the activation of the convolutional filter associated with this second texture for a SEM from the $0^{\circ}$ forged category.  This heat map shows where in the image this filter was most strongly activated--where the texture in the image matched that of the filter most closely.  As this figure shows, the activation is strongest in those regions of the image with diagonal structures.  

\begin{figure*}[t]
\begin{center}
\includegraphics [width=100mm]{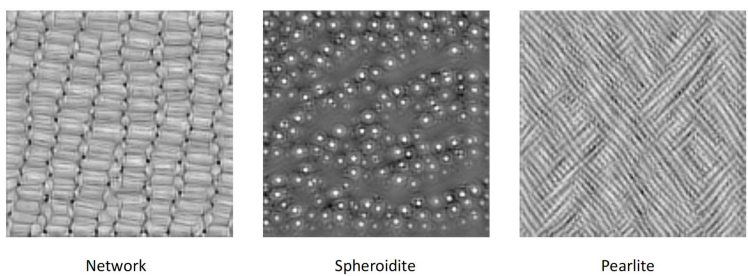}
\end{center}
\caption{Characteristic textures for each of the three categories in the steel data set: network, spheroidite and pearlite}
\label{texture_steel_fig} 
\end{figure*}

It is also possible to visualize which textures were most characteristic of each category.  These are the textures for which there is the greatest difference in intensity between classes.  Figure~\ref{texture_steel_fig} shows textures that were characteristic for each category in the steel case.  These characteristic textures can be compared to the SEM images of these classes in Fig.~\ref{steel_sem_fig}.  The characteristic texture for the network class mimics the cell-like structures in this microconstituent.  The characteristic texture for spheroidite is the same as the most important texture for the random forest classifier in this case, representing the dotted structure of spheroidite.  For the pearlite microconstituent, the characteristic texture resembles long, thin lamellar structures.  

\begin{figure*}[t]
\begin{center}
\includegraphics [width=90mm]{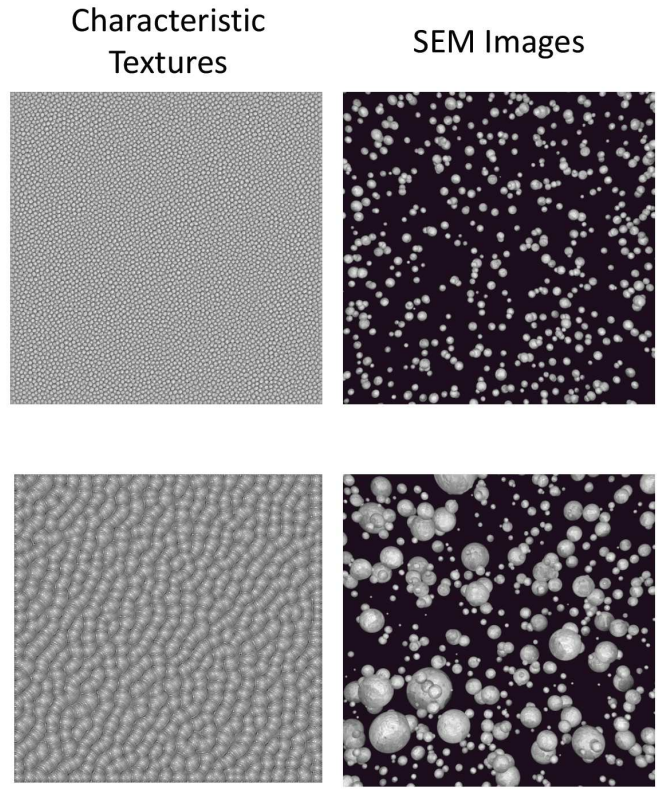}
\end{center}
\caption{Characteristic textures for two different particle distributions in the powder data set.  SEMs from each category are shown on top, and the corresponding characteristic textures are shown underneath.}
\label{texture_powder_fig} 
\end{figure*}

Figure~\ref{texture_powder_fig} shows two different characteristic textures for the powder data set: one for a particle distribution dominated by smaller particles, and another for a particle distribution with some larger particles. The characteristic texture for the distribution with mostly small particles resembles many adjacent small circles.  The characteristic texture for the distribution with some larger particles resembles overlapping larger circles.  These characteristic textures therefore clearly reflect the difference in particle size distribution.  

The ability to visualize the texture features is not only interesting, it is also useful.  The visualizations could be used to determine what particular textures are responsible for differences in mechanical properties.  While none of the data sets explored in this study included mechanical properties (due to the scarcity of public data sets that have both microstructure and mechanical properties available), it would be straightforward to use the texture features to not just classify the microstructure but to go one step further and predict the mechanical properties.  In those cases, it would be quite relevant to know which textures are most highly correlated with desirable mechanical properties.  Further work in this direction will be the focus of future work.

\section{Conclusions} \label{ConclusionsSec}

A key goal of this study was determining a method of featurizing SEMs that generalizes well between different data sets and prediction tasks.  Four different featurization strategies were evaluated.  While different featurization strategies exhibited the best performance in different test cases, the mean texture featurization had good overall performance.  Furthermore, this featurization is quite straightforward and generalizable.  It results in $N_{filters}$ features, which is far fewer features than are generated by either the Gram ($N_{filters}^{2}$) or VLAD ($N_{filters}\cdot N_{words}$) featurizations.  This smaller number of features is preferable because of its reduced memory requirements and because it allows the random forest classifier to be more computationally efficient, since the computational cost of this classification algorithm scales with the number of features.  Unlike the VLAD featurization, the mean texture featurization does not require any updates to a dictionary when new images are added to the data set, and the addition of new images does not affect the featurization of previous images.  

The performance of the random forest classifier was evaluated when using the outputs from five different convolutional layers.  This study demonstrated that the classifier performance was strongly dependent on which convolutional layer the texture features were derived from.  In the powder case, features derived from earlier convolutional layers were most effective.  In the steel case, on the other hand, the features from the later convolutional layers provided greater classification accuracy.  This difference can be resolved, it seems, by using the union of the features from all of the layers.  

Lastly, we investigated the interpretability of the texture features generated by the mean texture featurization strategy.  We showed that is possible to visualize these textures using an optimization method to determine what input image causes a given convolutional filter to be most strongly activated.  For each of the test cases, the most important texture features were determined.  For the steel test case, the characteristic textures for each category were presented.  These characteristic textures can be used to determine what textures are most differentiating about each microstructure.  While deep learning methods are often criticized as black box methods, this texture visualization method, which has not been previously applied in the context of material microstructures, can be used to gain insights into what differentiates various microstructures and how machine learning models built on these features are making their predictions.  

A key next step will be evaluating these featurization techniques on data sets for which material properties are also available.  In the three case studies presented in this paper, the goal of the machine learning model was to classify the SEM image.  However, it would be even more useful to be able to use the SEM image to make a prediction about the material properties.  Since microstructure is known to play a key role in macroscopic material properties, it would be logical to use the SEM texture features, along with information about composition and processing, to make predictions about the properties.  One missing link in this future work is the availability of multiple sizable data sets for which both SEM images and material properties are available.  While many such data sets surely exist, they have, for the most part, not been made publicly available.  The availability of such data sets would mark a key step forward in the development of scalable, generalizable SEM featurization techniques for use in building data-driven process-structure-property maps.  In this context, the visualization techniques presented in this paper could be applied to determine the textures that are characteristic of favorable mechanical properties.



\bibliographystyle{elsarticle-num} 
\bibliography{biblio}

\end{document}